%% file: main.tex
\documentclass{article}
\usepackage[utf8]{inputenc}
\usepackage{graphics}
\usepackage{graphicx}

\usepackage{hyperref}
\usepackage[margin=0.8in]{geometry}
\usepackage{array}
\usepackage{tabularx}
\usepackage{amsmath}
\usepackage{float}
\usepackage{authblk}
\usepackage{amssymb}

\usepackage[sorting = none, backend = bibtex, style=numeric-comp]{biblatex}

\usepackage{bm}
\usepackage[linesnumbered,ruled]{algorithm2e}
\usepackage{algpseudocode}
\usepackage{subfig}

\newcommand{\B}{\ensuremath{\mathbb{B}}}

\algrenewcommand\algorithmicrequire{\textbf{Input:}}
\algrenewcommand\algorithmicensure{\textbf{Output:}}

\title{Sampling binary sparse coding QUBO models using a spiking neuromorphic processor}

\author[1]{Kyle Henke\thanks{Email: khenke@lanl.gov}}
\author[1]{Elijah Pelofske\thanks{Email: epelofske@lanl.gov}}
\author[2]{Georg Hahn\thanks{Email: ghahn@hsph.harvard.edu}}
\author[1]{Garrett T. Kenyon\thanks{Email: gkenyon@lanl.gov}}

\affil[1]{Los Alamos National Laboratory, CCS-3 Information Sciences}
\affil[2]{Harvard University, T.H.\ Chan School of Public Health}

\begin{document}
\date{\vspace{-7ex}}

\maketitle

\begin{abstract}
\input{abstract}
\end{abstract}

\input{text}

\setlength\bibitemsep{0pt}
\printbibliography

\end{document}

%% file: abstract.tex
We consider the problem of computing a sparse binary representation of an image. To be precise, given an image and an overcomplete, non-orthonormal basis, we aim to find a sparse binary vector indicating the minimal set of basis vectors that when added together best reconstruct the given input. We formulate this problem with an $L_2$ loss on the reconstruction error, and an $L_0$ (or, equivalently, an $L_1$) loss on the binary vector enforcing sparsity. This yields a so-called Quadratic Unconstrained Binary Optimization (QUBO) problem, whose solution is generally NP-hard to find. The contribution of this work is twofold. First, the method of unsupervised and unnormalized dictionary feature learning for a desired sparsity level to best match the data is presented. Second, the binary sparse coding problem is then solved on the Loihi~1 neuromorphic chip by the use of stochastic networks of neurons to traverse the non-convex energy landscape. The solutions are benchmarked against the classical heuristic simulated annealing. We demonstrate neuromorphic computing is suitable for sampling low energy solutions of binary sparse coding QUBO models, and although Loihi~1 is capable of sampling very sparse solutions of the QUBO models, there needs to be improvement in the implementation in order to be competitive with simulated annealing. 

%% file: text.tex
\section{Introduction}
\label{section:introduction}
We are interested in the computation of a sparse binary reconstruction of an image. This task plays a role whenever an image of interest is not directly observable and instead must reconstructed from a limited sample or projection using compressive sensing. Sparse binary reconstruction is of interest in, for instance, the fields of radioastronomy and molecular imaging, as well as image compression \cite{Ting2006, Mohideen2021}. Sparse binary coding falls into the class of Quadratic Unconstrained Binary Optimization (QUBO). QUBO models are challenging computational problems that are difficult to solve exactly using classical algorithms due to exponential run time complexity, in general. QUBO models are a specific type of discrete combinatorial optimization problems, and in general it is of considerable interest to be able to compute optimal solutions of QUBO models more efficiently than existing methods. Networks of spiking neurons with noise have been shown to offer new opportunities for solving these problems. By programming the constraints into the architecture of a network of spiking neurons and controlling the frequency of network states during the resulting stochastic dynamics of the network, the exploration of complicated energy (e.g., objective function) landscapes describing our problem of interest can be performed in practical time.

Mathematically, given a signal $\boldsymbol{x} \in \mathbb{R}^m$ and an overcomplete and non-orthonormal basis of $n > m$ vectors $\boldsymbol{D} = \{ D_1,\ldots,D_n \}$, we aim to infer a sparse representation of the input using few elements from the dictionary. Here, an overcomplete set is defined as one that contains more functions than needed for a basis. All basis matrices as well as the image $\boldsymbol{x}$ are assumed to be of equal dimensions. The task is to find the minimal set of non-zero activation coefficients $\boldsymbol{a}$ that accurately reconstruct the given input signal $\boldsymbol{x}$, where $\boldsymbol{a} \in \mathbb{B}^n$ is a binary vector of length $n$ for $\mathbb{B}=\{0,1\}$. We can express the computation of a sparse binary representation of the image $\boldsymbol{x}$ using the basis $\boldsymbol{D}$ as the minimization of the energy function
\begin{equation}
    E \left( \boldsymbol{x}, \boldsymbol{a} \right) = \min_{\boldsymbol{a}} \left[ \frac{1}{2} \Vert \boldsymbol{x} - \boldsymbol{D} \boldsymbol{a} \Vert_2^2 + \lambda \Vert \boldsymbol{a} \Vert_0 \right]
    \label{eq:objective}
\end{equation}
where $\Vert \cdot \Vert_2$ is the Euclidean norm and $\Vert \cdot \Vert_0$ denotes the number of nonzero elements. The parameter $\lambda>0$ is a Lasso-type parameter \cite{Tibshirani1996} controlling the sparseness of the solution. A large value of $\lambda$ results in a more sparse solution to eq.~\eqref{eq:objective}, while smaller values yield denser solutions. Therefore, the parameter $\lambda$ allows one to effectively balance the reconstruction error (the $L_2$ norm) and the number of non-zero activation coefficients (the $L_0$ norm). Since eq.~\eqref{eq:objective} belongs to the class of 0-1 integer programming problems, finding a sparse representation falls into an NP-hard complexity class. The objective function of eq.~\eqref{eq:objective} is non-convex and typically contains multiple local minima.

We investigate a spiking neuromorphic processor to solve the binary sparse representation problem given by the objective function in eq.~\eqref{eq:objective}. Neuromorphic computing is a proposed computing model inspired by the human brain, which is able to complete learning tasks better than classical von Neumann computers \cite{roy2019towards, Davies2021, Schuman2022}.

\begin{figure*}[ht]
    \centering
    \includegraphics[width=.83\textwidth]{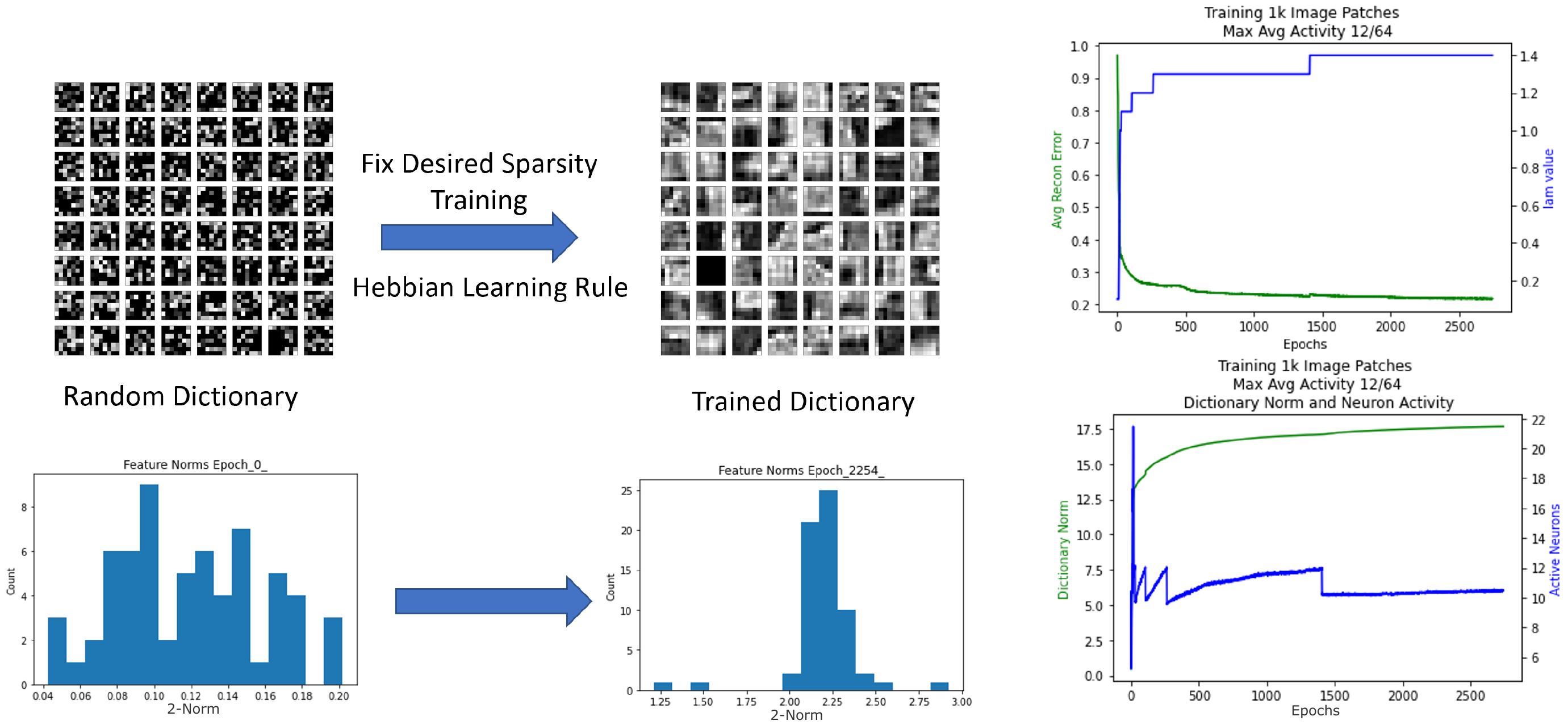}
    \caption{Randomly initialized dictionary with norms distributed between $.01$ and $.2$. After the training algorithm, norms increase and an optimal binary dictionary is learned for a fixed average activity of $12$ features. \label{fig:learning}}
\end{figure*}

\section{Methods}
\label{section:methods}

\subsection{Transformation relations}
\label{section:methods_transformation}
\begin{figure}[h]
    \centering
    \includegraphics[width=0.35\textwidth]{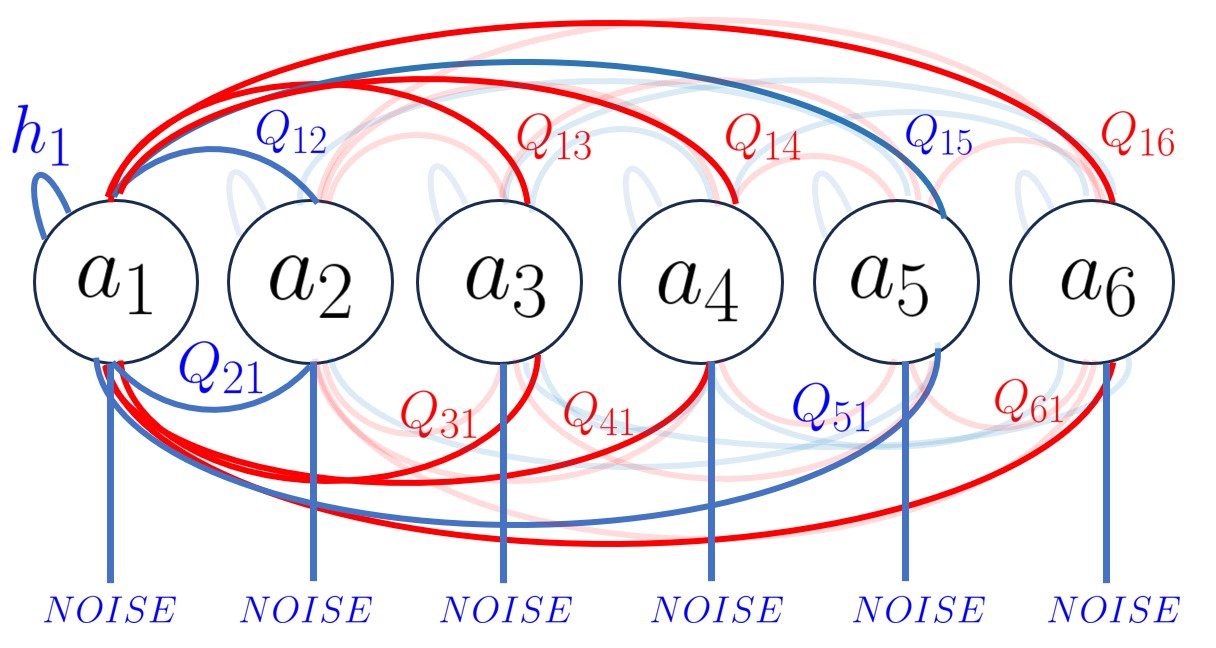}
    \caption{Network connectivity of the variables in eq.~\eqref{eq:H}. Connections include the self interaction terms $h_i$ (symmetric weights proportional to the inner product between features), the inter-neuron connection weights $Q_{ij}$, and the stochastic noise input. Red is inhibitory connection and blue is excitatory. Network is sampled at different times and activity is measured for solution.}
    \label{fig:NetworkDynamics}
\end{figure}

The problem being solved has to be given as a QUBO problem. In this formulation, the observable states of any neuron is $0$ and $1$. We start by reformulating eq.~\eqref{eq:objective} in QUBO form. To this end, we observe that for $\boldsymbol{a} \in \B^n$,
\begin{align*}
    E(\boldsymbol{a}) &= \frac{1}{2} \Vert \boldsymbol{x} - D \boldsymbol{a} \Vert_2^2 + \lambda \Vert \boldsymbol{a} \Vert_0\\
    &= \frac{1}{2} \boldsymbol{x}^\top \boldsymbol{x} - \boldsymbol{x}^\top D \boldsymbol{a} + \frac{1}{2} \boldsymbol{a}^\top D^\top D \boldsymbol{a} + \lambda \sum_{i=1}^n a_i.
\end{align*}
As expected, multiplying out eq.~\eqref{eq:objective} yields a quadratic form in $\boldsymbol{a}$, meaning that we can recast our objective function as a QUBO problem. For this we define the following two transformations:
\begin{align}
    h_i = -D_i^\top \boldsymbol{x} + \lambda + \frac{1}{2} D_i^\top D_i,\qquad
    Q = \frac{1}{2} (D^\top D).
    \label{eq:hQ}
\end{align}
Using eq.~\eqref{eq:hQ}, we can rewrite eq.~\eqref{eq:objective} as a QUBO, given by
\begin{align}
	H(\boldsymbol{h}, Q, \boldsymbol{a}) = \sum_{i=1}^n {h_i a_i} + \sum_{i<j} Q_{ij} a_i a_j,
    \label{eq:H}
\end{align}
which is now in suitable form to be solved on Intel's Loihi neuromorphic chip \cite{Henke2020alien,Henke2020machine}. Network connectivity mapping can be seen in Figure~\ref{fig:NetworkDynamics}, where $a_i$ denote the neurons, $h_i$ are the self interactions on the neurons, and $Q_{ij}$ are the inter-neuron connection weights.

\subsection{Loihi neuromorphic chip implementation}
\label{section:methods_loihi_1_chip_implementation}
Intel's Loihi 1 is the first generation neuromorphic computing device that draws inspiration from biology to implement spiking neural networks with neurons as the fundamental processing elements \cite{Davies2021survey}.

\begin{figure}[h!]
    \centering
    \includegraphics[width=0.35\textwidth]{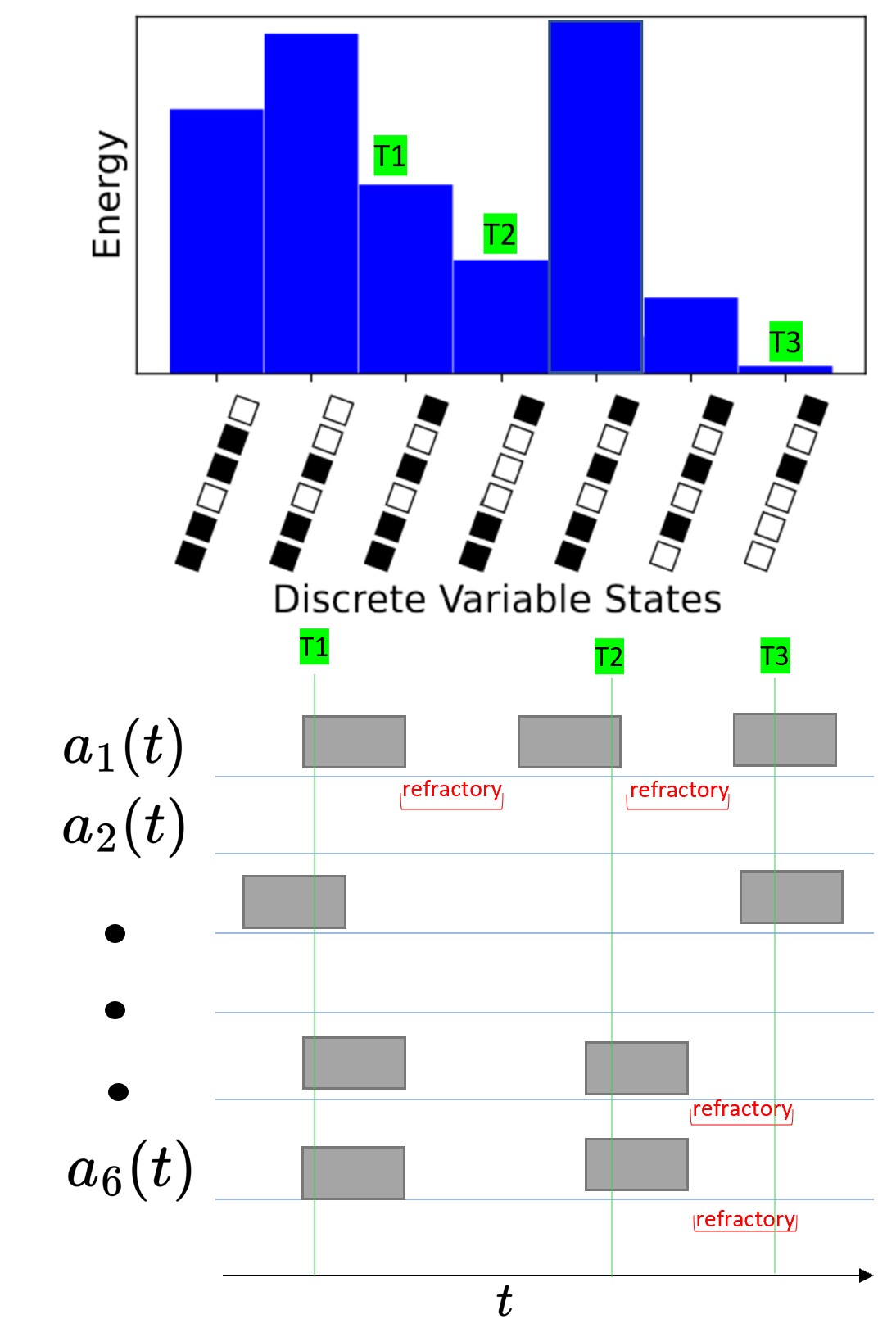}
    \caption{Conceptual diagram of how we expect spike-based dynamics support the bypassing of high-energy barriers. Energy, e.g., the objective function evaluation for a set of variable assignments, is given on the y-axis and the x-axis shows variable assignments where $\blacksquare$ denotes $+1$ and $\square$ denotes $0$ (for the chosen number of variables of $n=6$). In this example, the relatively sparse state of $(0, 0, 0, 1, 0, 1)$ has the lowest overall energy. When the system is sampled at different time periods T1, T2, and T3, we are able to bypass the largest energy barrier because the refractory period automatically shuts off variables 5 and 6 \cite{Jonke2016}.}
    \label{fig:EnergyTime}
\end{figure}

\subsubsection{Overcoming local minima on Loihi~1}
\label{section:methods_overcoming_local_minima}

Compared to a Boltzmann machine \cite{hinton2007boltzmann}, spiking networks allow for transitions between extreme objective function variable states (see Figure~\ref{fig:EnergyTime}). Because of the limited time of activity, or forced refractory period, defined by $\tau$, active neurons are turned off for a determined time and others who were inhibited by the active neuron now have a chance to activate. These periods allow the network to explore non-locally and facilitate the bypassing of high energy barriers in the optimization landscape \cite{Guerra/fnins.2017.00714,Jonke2016}. After the refractory period is over, previously active neurons will likely re-fire because they are receiving a strong input and a low-energy state will again be found. Figure~\ref{fig:EnergyTime} demonstrates this property through the substantial variation in the energy reads obtained from Loihi~1 as a function of time. High energy read outs correspond to refractory periods of neurons active in the ideal solution, and the repeated lowest energy reflects the return to lower energy solution states \cite{Davies2021survey}. For the QuboSolver method ran on Loihi~1, a threshold mantissa of 96, weight exponent of 6, and noise mantissa of 0 and exponent of 7 are used. In order to sample each QUBO on Loihi~1, a total of $2,000$ samples are measured; $4$ simulation times ($5,000, 10,000, 15,000, 20,000$) are varied over, and $5$ different weight matrix scalings ($10, 100, 1000, 10000, 100000$) are varied, with each parameter combination being sampled $100$ times (this gives $4 \cdot 5 \cdot 100 = 2000$ samples per QUBO).

\subsubsection{Un-normalized Dictionary Learning}
\label{section:methods_un_normalized_dictionary_learning}
Sparse coding optimization can be seen as a two step process where a dictionary is first learned in an unsupervised way by using a local Hebbian rule. Typically, when learning a basis for solving the convex Lasso problem, the algorithm requires the re-normalization of the columns of the dictionary $D$ after each learning epoch. The normalization is critical for convergence in the Lasso setting because the values of the sparse vector $\boldsymbol{a}$ are allowed to take on any value. Previous work has demonstrated the ability to learn a dictionary in a QUBO regime, but this required the introduction of a new amplification parameter $\beta$ to the input \cite{Henke2020alien,Henke2020machine}. Here, we introduce a new learning technique that allows the algorithm to find the optimal norm for features based up on a predetermined desired average level of sparsity defined as $\boldsymbol{s}\in (0,1)$. The dictionary is initialized with features drawn from a normal distribution with random norms below 1 and a small sparsity penalty parameter $\lambda$. After solving the binary sparse coding problem for each sample in the training data, the dictionary is updated. If the average sparsity over the training epoch is above the desired level $\bf{s}$, the penalty parameter $\lambda$ is increased for the next epoch. Pseudo code for the algorithm is presented below and the learning results are summarized in Figure \ref{fig:learning}. We can see the average neuron activity and reconstruction error converge along with the norms of the learned features.

We applied our technique to a patched version of the standard fashion MNIST (fMNIST) data set \cite{xiao2017fashionmnist}. Each 28x28 image was broken up into 16 7x7 patches and we selected a dictionary of size $64$ in order to partition the problem into sub-problems which could be implemented on Loihi~1 (the exact number of variables for the sub-problems is arbitrary but fixed). Even with a smaller data structure, it was still necessary to perform our dictionary learning algorithm using the classical simulated annealing approach when solving for our sparse code in step~6 of Algorithm~\ref{alg:dict_update_sparsity}. The Lasso parameter $\lambda$ was increased from $0.1$ to $1.4$ in increments of $0.1$ to adapt to the sparsity of the solution (see the top right plot in Figure~\ref{fig:learning}).

\begin{algorithm}[h!]
    \caption{Dictionary Update}
    \label{alg:dict_update_sparsity}
    \SetKwInOut{Input}{input}
    \SetKwFor{For}{for}{}{end}
    \SetKwFor{Function}{function}{}{end}
    \Input{$\bm{D} \in \mathbb{R}^{m \times n}$, $Train\_data \in \mathbb{R}^{b \times m}$, $\eta \in \mathbb{R}^+$, $\bm{s} \in (0,1)$, $\lambda>0$, \text{number of epochs} $N$}
    \Function{\textnormal{learn\_dictionary}($D$, $a$, $x$, $\eta$, $s$,\textnormal{number of epochs})}{
        \For{$\textnormal{epoch} = 1,2,\ldots,N$}{
            $activity\_count = 0$\;
            \For{$i = 1, 2,...,b$}{
                $x = Train\_data[i]$\\
                $\textnormal{Solve for}~a$\\
                $recon = Da$\\
                $residual = x - recon$\\
                $\Delta D = residual \hspace{1mm} a^T$\\
                $D = D + \eta \Delta D$\\
                $activity\_count = activity\_count + sum(a)$
            }
            \If{$\frac{activity\_count}{n*b}>\bm{s}$}{
                $\lambda = \lambda + 0.1$
            }
        }
    }
    \Return{$D$}
\end{algorithm}

\section{Results}
\label{section:results}
Figure~\ref{fig:learning} visualizes the successful implementation of un-normalized dictionary feature learning. Using a local learning rule and a fixed sparsity level, we can see that the algorithm learns a better basis for reconstruction as the average error of the training data decreases over training epochs and it also converges to the desired average sparsity level.

\begin{figure}[h]
    \centering
    \includegraphics[width=0.4\textwidth]{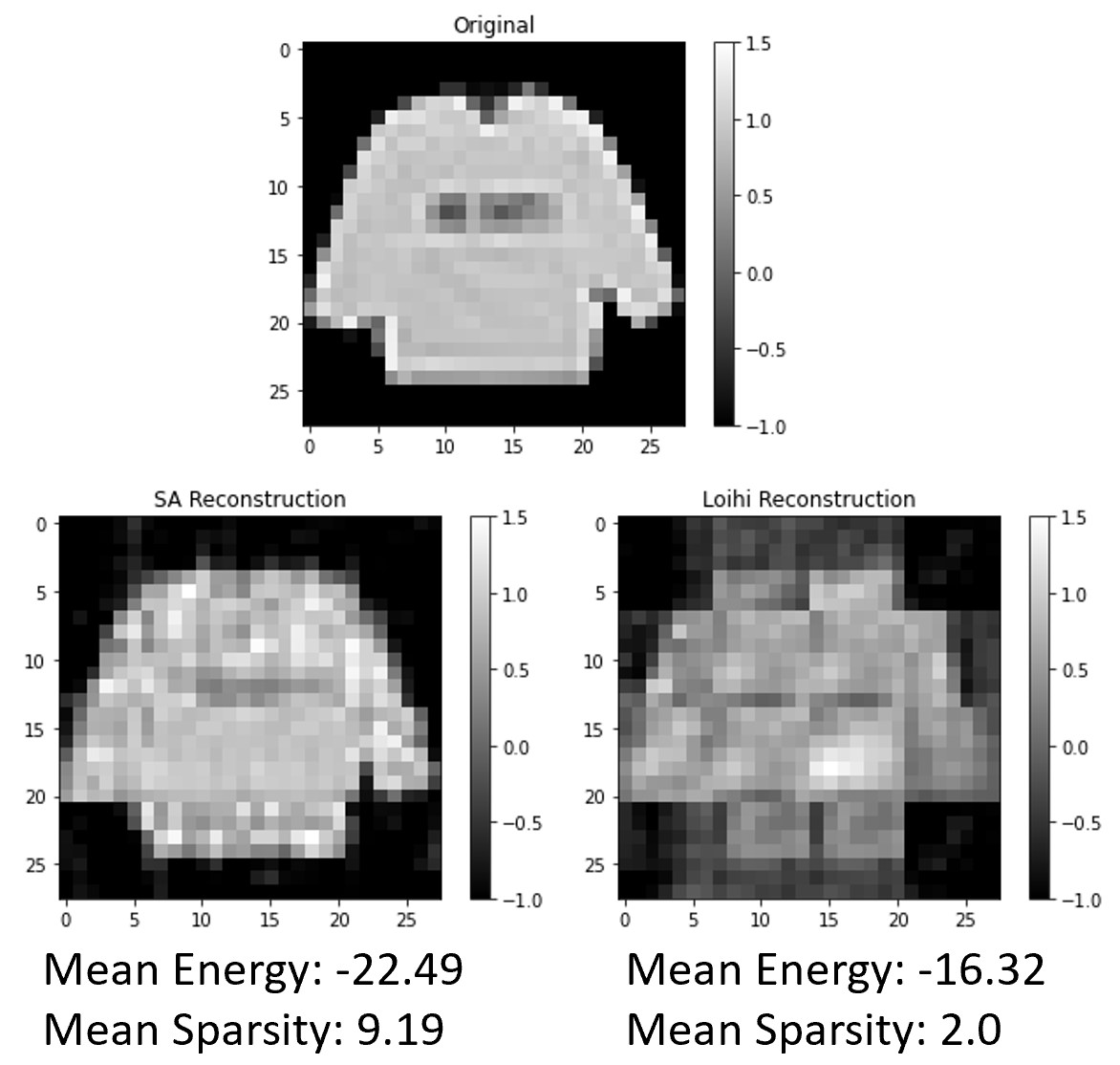}
    \caption{Reconstructions from classical SA and Loihi~1. Full image consists of 16 separate QUBO solves and the mean energies and sparsity levels are displayed. The sparsity levels are the mean (across the $16$ QUBO models) number of variables in the lowest energy state which were in the state of $+1$. }
    \label{fig:reconstructions}
\end{figure}

After successfully training each dictionary with simulated annealing (SA), a total of $16$ separate QUBO models are generated. Each QUBO is then sampled using Loihi~1 (see Section \ref{section:methods_loihi_1_chip_implementation}). In order to provide a reasonable comparison against existing classical heuristic algorithms, we also sample each of the $16$ QUBO models using simulated annealing. The simulated annealing implementation we use is a D-Wave SDK implementation \cite{sa}, using $1000$ samples per QUBO and all default settings. Using the best solutions (e.g., the computed variable assignments with the lowest energy found among all samples) from both Loihi~1 and simulated annealing, we can reconstruct the original image from sampling all $16$ QUBOs. These reconstructions are shown in Figure \ref{fig:reconstructions}. Although SA has a lower mean energy, Loihi~1 is able to find reasonable solutions at much lower average sparsity levels. Similar to previous demonstrations of lower power usage for certain applications \cite{Davies2021,Henke2022,Henke2020alien,Henke2020machine,Fast}, Loihi~1 uses an average power consumption of $\sim 0.0192$ joules per sample, per QUBO matrix compared to an average power consumption of $\sim 0.115$ joules per sample per QUBO matrix for simulated annealing. The simulated annealing power consumption was measured using pyRAPL \footnote{\url{https://pyrapl.readthedocs.io/en/latest/}} (including RAM power usage). The total power usage was computed by subtracting the idle machine power consumption (for the same time duration) from the power consumption when simulated annealing was run. The Loihi~1 power consumption was measured using the nxsdk power monitoring function.

\begin{figure}[h]
    \centering
    \includegraphics[width=0.4\textwidth]{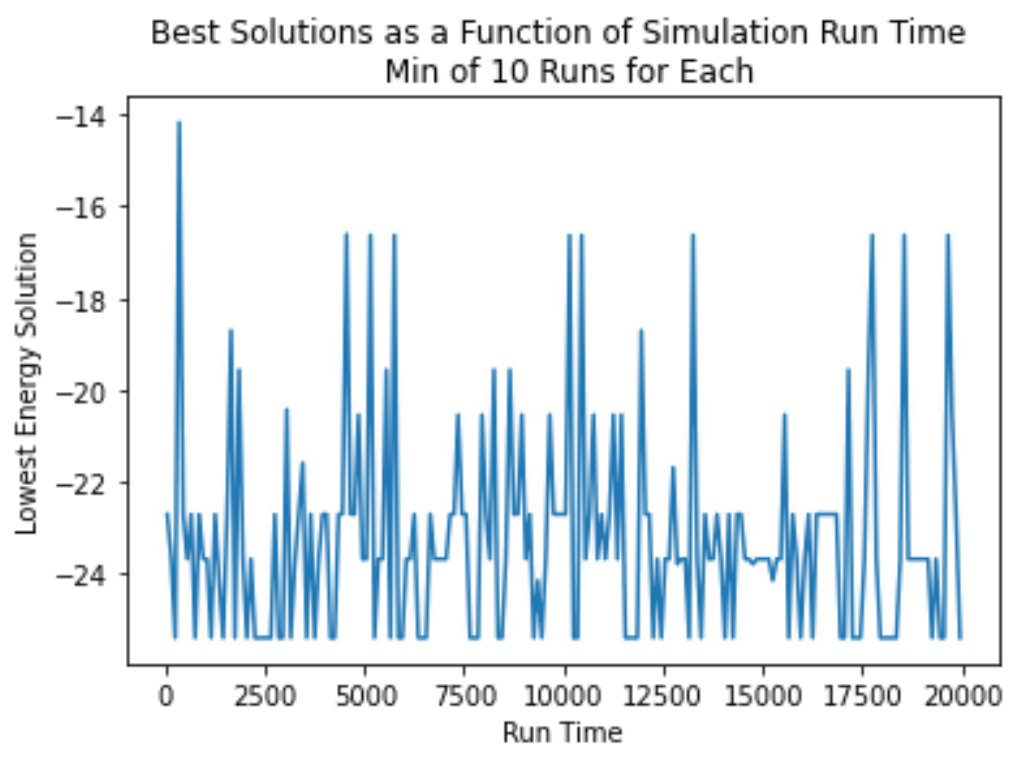}
    \caption{QUBO energies read out at different simulation times (minimum of $10$ readouts per simulation time) from the Loihi~1 neuromorphic processor for a single QUBO patch.}
    \label{fig:energies}
\end{figure}

\section{Discussion and Conclusion}
\label{section:conclusion}
In this work, we derived a technique for learning an unmormalized dictionary for binary sparse coding in an unsupervised manner when given a desired sparsity level. The trained dictionary was then used for solving the binary sparse coding problem in the form of a QUBO using the Loihi~1 spiking neuromorphic processor and compared against simulated annealing. Measurements taken from Loihi~1 demonstrate the use of refractory periods and stochasticity allow the spiking processors to overcome large energy barriers in the non-convex landscape. The solutions from Loihi~1 are not of the same quality compared with simulated annealing, but it is interesting to note that the solutions are considerably sparser, and use less energy to compute each sample compared to simulated annealing. 

Future work could include comparing the results on Loihi~2, the second generation of Intel's spiking processor. Using an iterative warm start approach with Loihi, where the best solution found at each iteration is used to initialize the system at the next iteration, similar to an iterative warm start algorithm in classical optimization, could improve the total space explored and thus the likelihood of finding a global minimum.

\section{Acknowledgements}
This work was supported by the U.S.\ Department of Energy through the Los Alamos National Laboratory. Los Alamos National Laboratory is operated by Triad National Security, LLC, for the National Nuclear Security Administration of U.S.\ Department of Energy (with Contract No.~89233218CNA000001). We gratefully acknowledge support from the Advanced Scientific Computing Research (ASCR) program office in the Department of Energy's (DOE) Office of Science, award \#77902 along with funding from the NNSA's Advanced Simulation and Computing Beyond Moore's Law Program at Los Alamos National Laboratory. This work has been assigned the technical report number LA-UR-23-25877.